\documentclass[lettersize,journal]{IEEEtran}
\usepackage{amsmath,amsfonts}
\usepackage{algorithmic}
\usepackage{algorithm}
\usepackage{array}
\usepackage[caption=false,font=normalsize,labelfont=sf,textfont=sf]{subfig}
\usepackage{textcomp}
\usepackage{stfloats}
\usepackage{url}
\usepackage{verbatim}
\usepackage{graphicx}
\usepackage{cite}
\usepackage[table]{xcolor}

\usepackage{makecell}
\newcommand{\PreserveBackslash}[1]{\let\temp=\\#1\let\\=\temp}
\newcolumntype{C}[1]{>{\PreserveBackslash\centering}p{#1}}

\begin{document}

\title{Lightweight Regression Model with Prediction Interval Estimation for Computer Vision-based Winter Road Surface Condition Monitoring}

\author{Risto Ojala, Alvari Seppänen 
    \thanks{Related code is available at https://github.com/ojalar/SIWNet \newline 
    DOI: 10.1109/TIV.2024.3371104 \newline
    This work is licensed under a Creative Commons Attribution 4.0 License. For more information, see https://creativecommons.org/licenses/by/4.0/}
    }




\maketitle

\begin{abstract}
Winter conditions pose several challenges for automated driving applications.
A key challenge during winter is accurate assessment of road surface condition, as its impact on friction is a critical parameter for safely and reliably controlling a vehicle.
This paper proposes a deep learning regression model, SIWNet, capable of estimating road surface friction properties from camera images.
SIWNet extends state of the art by including an uncertainty estimation mechanism in the architecture.
This is achieved by including an additional head in the network, which estimates a prediction interval.
The prediction interval head is trained with a maximum likelihood loss function.
The model was trained and tested with the SeeingThroughFog dataset, which features corresponding road friction sensor readings and images from an instrumented vehicle.
Acquired results highlight the functionality of the prediction interval estimation of SIWNet, while the network also achieved similar point estimate accuracy as the previous state of the art.
Furthermore, the SIWNet architecture offers a more favourable balance of accuracy and computational load than previous state-of-the-art models.
\end{abstract}


\section{Introduction}
\IEEEPARstart{F}{riction} between the road and vehicle tyres plays a key role in defining how a vehicle should be controlled and manoeuvred in winter conditions.
The impacts of friction on driving safety are substantial, directly affecting factors such as braking distance and slip angle.
The friction between a vehicle tyre and the road depends on both the tyre and road surface properties.
Whereas the tyre properties remain mostly static, the road surface properties can greatly vary, especially during winter in countries with below-freezing temperatures.
It has been explicitly noted in the literature that accident rates are strongly affected by road surface condition \cite{wallman2001friction}.
In winter conditions, the road surface friction properties are mainly dependent on the amount of snow, ice, and water on the road.
Quantifying these road friction properties is essential for modelling, estimating, and predicting the friction between the tyre and the road.
Consequently, vehicle control needs to adapt to the prevailing road surface conditions to ensure safe operation.
As the availability and development of automated driving features is on the rise, the estimation methods for road surface friction properties have an increasingly important role.
Automated driving solutions must be capable of adapting to different friction conditions, modifying their control based on the environment.

Road surface condition can be analysed with several on-board methods \cite{acosta2017road, ma2022current}.
A common approach has been to utilise wheel dynamics information for the estimation task.
However, these methods generally have difficulties quantifying the road surface friction properties before severe slippage has occurred.
Special optical sensors have also been developed for analysing road surface condition, yet these are generally too expensive for consumer vehicle applications.
Recently, computer vision solutions with typical visible light cameras have been a popular approach for the task.
Computer vision approaches offer the convenience of utilising the existing windshield camera equipment.
Additionally, they have the potential of assessing the road surface condition in front of the vehicle, which enables predictive actions.

This paper expands computer vision-based road surface condition monitoring in winter conditions by presenting a deep learning model, SIWNet, for the task.
The regression model was developed to predict a scalar estimate for the road surface friction properties, summarising the effects of visible snow, ice, and water on tyre-road friction.
Hence the name of the model, SIWNet (Snow, Ice, Water Network).
Additionally, the model estimates the uncertainty of the prediction by providing a prediction interval with a multi-head architecture.
Such feature has not been previously proposed for regression-based road surface condition monitoring models.
Furthermore, SIWNet has been designed with practical on-board deployment in mind, with the model being computationally lightweight, yet achieving similar accuracy as previous state of the art.
Similarly to \cite{vosahlik2021self}, SIWNet was trained in an automated manner by matching images and corresponding friction information.
However, the work presented here utilised the SeeingThroughFog-dataset \cite{bijelic2020seeing}, where the friction values have been acquired with an optical road friction sensor.

The novel contributions of this work can be summarised as:
\begin{itemize}
  \item[$\bullet$]  SIWNet is the first road surface friction regression model to feature prediction intervals in the estimates.
  \item[$\bullet$]  SIWNet is computationally lightweight, with better accuracy and computational load trade-off than previous models.
  \item[$\bullet$]  This paper is the first work training a road surface friction estimation computer vision model for winter conditions based on optical road friction sensor data.
\end{itemize}

\newpage

\section{Related Work}
\subsection{Road surface condition monitoring}
Road surface condition monitoring is essential for different vehicle applications, and the topic has been studied in the past with several approaches.
Current interest in automated driving solutions has increased the importance of the field, as vehicle control algorithms require information of road surface condition to properly assess the situation.
Road surface condition monitoring is especially important in winter conditions, where the road surface friction properties greatly vary.
The existing methods can be divided to contact and non-contact based methods.

Contact methods are not analysed here in-detail, as this work is focused on computer vision methodology.
For a more thorough overview of contact-based methods, the review by Acosta \textit{et al.} \cite{acosta2017road} is recommended.
Fundamentals of contact-based estimation methods are still briefly presented here.
Contact-based methods generally measure the actual friction between the tyre and the road.
Road surface condition can be consequently estimated from this information.

Of the contact-based methods, slip-based methods are the most common. 
Measuring wheel rotation information and inertial measurement unit readings, friction between the tyre and the road can be estimated \cite{gustafsson1997slip}.
Utilising this information, the road surface condition can be defined based on the friction characteristics.
In a commercial vehicle, this information is readily available in the anti-lock braking unit and electronic stability control.
However, slip-based methods typically cannot accurately estimate the road surface conditions until severe slippage has already occured.
The review by Acosta \textit{et al.} \cite{acosta2017road} notes that slip-based approaches are typically considered inadequate for reliably improving the functionality of advanced driver assistance systems.
Other contact-based methods for road surface monitoring are based on vibration.
Low frequency methods can utilise signals such as the vehicle rotation speed \cite{pavkovic2006experimental}.
However, the approaches rely largely on slip-slope assumptions, and consequently lack robustness.
High frequency methods have yielded impressive results, yet these methods rely on additional sensor equipment.
Works on the topic have utilised microphones installed on the vehicle to monitor the tyres \cite{alonso2014board}.
This might not be a commercially feasible solution due to the burden of additional sensor installations.

To enable alternative means for estimating road surface condition, a number of non-contact methods have been developed.
Non-contact methods typically exhibit different operation characteristics and attributes compared to the contact-based methods. 
They offer alternative ways for the measurement, or they could be fused with contact-based methods.
Review of non-contact friction estimation has been presented by Ma \textit{et al.} \cite{ma2022current}.
The non-contact approaches can be roughly divided into methods that utilise special optical sensors, and methods which utilise computer vision algorithms to process images captured with traditional visible light cameras.
Additionally, there have also been some studies exploring road surface condition monitoring with automotive radars \cite{viikari2009road}.

With dedicated optical sensors, non-contact approaches commonly utilise infrared spectroscopy \cite{jonsson2012infrared}.
The approach is based on the different reflectance characteristics of water, ice, and snow.
This measurement approach is commonly utilised, and multiple commercial products applying the method are available on the market \cite{dsc111, dsc211, md30}.
Another optical technique for road surface condition monitoring is based on analysing polarisation of light \cite{casselgren2012polarization}.

Computer vision applied on regular visible light cameras has been a popular topic for non-contact road surface condition estimation.
The approach is lucrative from a practical point-of-view, since modern vehicles are equipped with forward-facing cameras.
Review of computer vision-based estimation has been prepared by Wu \textit{et al.} \cite{wu2020survey}.
Generally, computer vision approaches utilise machine learning models such as convolutional neural networks (CNNs) to analyse images of the road.

Several works utilising computer vision have applied classification techniques to assess the road surface condition.
Nolte \textit{et al.} \cite{nolte2018assessment} proposed applying CNNs to perform this classification task.
They trained ResNet50 \cite{he2016deep} and InceptionV3 \cite{szegedy2016rethinking} models to recognise six distinct categories of road surface: asphalt, dirt, grass, wet asphalt, cobblestone, and snow.
Similar studies have been conducted by Šabanovič \textit{et al.} \cite{vsabanovivc2020identification}, who developed a CNN capable of classifying the road pavement type (asphalt, cobblestone, gravel) as well as the surface condition (dry, wet).
The effectiveness of the developed road surface monitoring system was demonstrated with vehicle braking tests.
The tests highlighted that the stopping distance was shorter when utilising an adaptive anti-lock braking system control strategy, which was tuned based on the classification results.
To extend the capabilities of road surface classification, Wang \textit{et al.} \cite{wang2021road} proposed applying a segmentation CNN to perform the classification task.
Their classification task featured a total of nine different pavement and road surface condition combinations, including categories from winter conditions.

To further enhance the development of computer vision-based road surface monitoring in summer conditions, Cordes \textit{et al.} \cite{cordes2022roadsaw} published an open classification dataset called RoadSaw.
They collected the dataset in an automated manner, mounting an optical road surface monitoring sensor on a vehicle, as well as a forward-facing camera.
As a result, a large dataset of images in realistic road conditions was captured, along with the corresponding road surface condition readings from the sensor.
For further processing, the images were transformed to bird's-eye-view and differently sized patches (2.56 $m^2$, 7.84 $m^2$, 12.96 $m^2$) of the road  were cropped.
The optical road surface monitoring sensor readings were synchronised to the image patches by utilising timestamps and vehicle velocity information.
The dataset featured three different pavement types (asphalt, cobblestone, concrete), which had four different surface conditions (dry, damp, wet, very wet).
The authors evaluated the performance of a MobileNetV2 \cite{sandler2018mobilenetv2} classifier on the data.
They also proposed adding deterministic uncertainty quantification \cite{van2020uncertainty} functionality to the network, allowing assessment of the uncertainty of the classification predictions.

Classification-based road surface monitoring has also been extended to model the road surface in finer spatial detail.
In the camera view, several road surface conditions may be visible simultaneously, such as patches of snow on an asphalt road.
Providing a single classification for the entire visible road may result in inadequate assessment of the prevailing situation.
Roychowdhury \textit{et al.} \cite{roychowdhury2018machine} proposed a two-stage approach for detailed analysis of the road surface in winter conditions.
After classifying the overall road surface condition with a CNN, they split the image of the road into a grid, in which each cell was classified separately.

The resolution of road surface condition estimation has been also extended by applying regression models to the task, instead of utilising classification models.
In terms of friction-related information, regression models allow for more accurate representation of the road surface properties.
This is due the model predicting a continuous value, instead of utilising a discrete number of categories labelled with certain values.
Vosahlik \textit{et al.} \cite{vosahlik2021self} proposed automated training of a CNN regression model based on corresponding friction information derived from a slip-based contact method.
Based on data acquired with a 1:5 scale car model, they created a dataset of matching friction values and images, which included samples from winter conditions.
Their contact-based method estimated the friction between the road and the tyres, which was used to label the images.
From the images, they cropped out a region of 1.5x1.5 metres in front of the vehicle, which was transformed to bird's-eye-view for processing.
Known vehicle speed and timestamps were utilised to synchronise the friction readings and the image patches.
They trained a ResNet50 network on the data to perform the regression task.
Regression-based road surface friction estimation was also developed by Du \textit{et al.} \cite{du2023pavement}, who studied the problem in summer conditions.
They collected road surface condition information with a commercial contact-based monitoring system, which utilised two measurement wheels for estimating the tyre-road friction coefficient.
The measurement system was equipped on a truck, which was followed by a vehicle equipped with a camera gathering image data of the road.
The test route was divided into 50 metre long sections, in which the captured images were labelled with the mean of the friction readings within the specific section.
For image processing, smaller patches of the road were cropped from the road area between the regions through which the measurement wheels had traversed.
In their work, they trained different CNN regression models to predict the road surface friction.
The vision-based estimators were fused with another machine learning model, which processed the vehicle dynamics information.
The fusion model seemed to offer clear benefits, as several key performance metrics were improved.

\subsection{Prediction intervals for deep learning regression models}
Uncertainty assessment has been a key topic in deep learning due to the obscurity of the data-driven the models.
In order to evaluate the reliability of the estimates produced by deep learning models, several approaches have been proposed to quantify related uncertainties.
An extensive review on the topic has been written by Abdar \textit{et al.} \cite{abdar2021review}.
Most extensive uncertainty quantification methodologies include Bayesian neural networks and ensemble techniques.
Bayesian neural networks model the network weights as distributions, allowing the predictions to be accurately modelled as posterior distributions.
Ensemble models rely on utilising multiple neural network models to process the input, and determining the uncertainty based on the outputs of the networks.
Bayesian and ensemble methods have been proven to generally provide reliable uncertainty information.
However, these methods typically require immense computational resources to operate, and applying them is often not practically feasible.

A common approach for assessing uncertainty in regression problems is the estimation of prediction intervals \cite{dewolf2023valid}.
Several approaches have been proposed for generating prediction intervals with deep learning regression models, including the previously mentioned Bayesian and ensemble methods.
However, more lightweight approaches for estimating prediction intervals have also been developed.
Typically this has been achieved by modifying the neural network architecture to feature an additional output node.
In order to quantify uncertainty, Nix and Weigend \cite{nix1994} proposed adding a separate output node with its own hidden layers to a fully connected network architecture.
The additional output node was responsible for estimating the variance of each prediction on each forward pass, whereas the other output node remained responsible for producing the point estimate.
Modelling each network prediction as a probability distribution, the network could be trained with a maximum likelihood loss function.
In such architecture, uncertainty is quantified by the value of the variance output, which can also be used to generate prediction intervals.
Somewhat similar prediction interval estimation has been proposed by Khosravi \textit{et al.} \cite{khosravi2010lower}. 
Their approach was based on two output nodes, which were responsible for predicting the lower and upper bound of the prediction interval.
A special loss function was used for training, which determined the target coverage of the prediction interval.

\subsection{Research gap}
This paper aims to enhance the existing state of the art of winter road surface condition evaluation.
A novel CNN architecture, SIWNet, is proposed for the task of computer vision-based estimation of road friction properties.
Similarly to the work of Vosahlik \textit{et al.} \cite{vosahlik2021self}, SIWNet is implemented as a regression model.
SIWNet was trained and tested based on an optical road friction sensor ground truth.
Such approach has not been previously applied in the literature in winter conditions, and therefore this study presents unique results highlighting the functionality of the chosen approach.

SIWNet expands state of the art by including uncertainty quantification in the regression architecture.
This is achieved by including an additional prediction head, which enables representation of the model output as a prediction interval.
The approach is based on the work of Nix and Weigend \cite{nix1994}.
Uncertainty quantification is a vital feature for computer vision-based road surface monitoring systems, as visual estimation of road surface friction properties is bound to feature varying levels of uncertainty.
For example, some surface conditions may be difficult to detect with computer vision, or other road users may be partially blocking the view of the road.
Cordes \textit{et al.} \cite{cordes2022roadsaw} have previously proposed an uncertainty estimation approach in their classification-based work.
However, studies developing regression models for the prediction task have not proposed methods for uncertainty quantification.

In addition to including an uncertainty quantification mechanism, SIWNet is also designed to feature a computationally lightweight architecture.
The findings of this paper highlight that computer vision-based winter road surface friction regression is not dependent on an extensively large neural network architecture.
Previous research \cite{vosahlik2021self} on the topic has applied a relatively large model for the task.
This finding is therefore a notable improvement, considering the generally limited vehicle on-board computational resources.

\section{Methods}
\subsection{Problem formulation and dataset}
The goal of the research was to develop a model capable of accurately assessing the winter-related road surface friction properties based on images captured from a typical vehicle windshield camera.
Additionally, the work aimed to enhance the robustness of computer vision-based road friction regression by introducing uncertainty quantification methodology to mitigate the natural inaccuracies of visual road friction estimation.
Simultaneously, the developed model must be lightweight to allow for on-board deployment with limited computational hardware.
These goals were pursued by formatting the computer vision problem as a task of predicting a ground truth road friction value from a corresponding image of the road.

\begin{figure}
    \centering
    \includegraphics[width=0.45\textwidth]{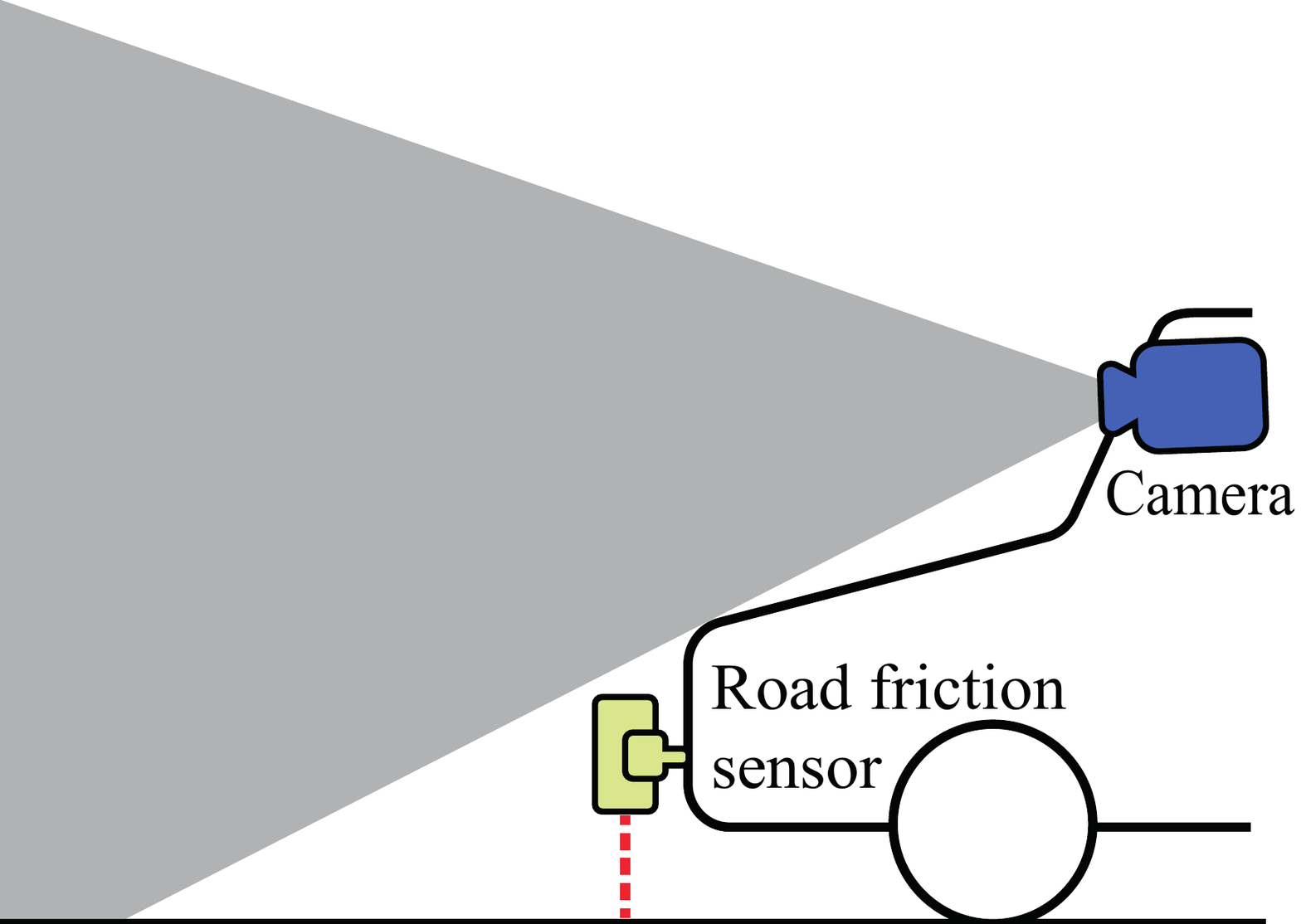}
    \caption{\small Illustration of the sensor installations, with the camera at the windshield and the road friction sensor at the bumper.}
    \label{fig:sensors}
\end{figure}

To acquire relevant data for the research, the publicly available SeeingThroughFog dataset \cite{bijelic2020seeing} was utilised.
The dataset has been gathered by driving an instrumented vehicle in central European and Nordic countries during winter.
Part of the dataset has also been recorded in the springtime, to include summer-like conditions.
The instrumented vehicle contained a forward-facing camera, as well as an optical road friction sensor.
The sensor setup is illustrated in Fig. \ref{fig:sensors}.
In this paper, the camera images and corresponding road friction sensor readings were utilised for developing a neural network model, SIWNet, for assessing the road surface condition.
SIWNet is trained to process an image of the road, and predict a value corresponding to the road friction sensor reading.

The optical friction measurement unit used in the dataset is a prototype from a widely recognised manufacturer, seemingly similar to those of the manufacturer's other models \cite{dsc111, dsc211, md30}.
Similarly to the commercial equivalents of the sensor, the unit measures the amount of snow, ice, and water on the road surface.
By manufacturer software, these road surface friction properties are summarised into a single factor, \textit{grip factor}, within range 0.09...0.82.
This factor effectively indicates the slipperiness of the road.
Here, this grip factor was normalised to a range 0.00...1.00, and called \textit{friction factor}, $f$.
It should be noted that the friction factor is not the friction coefficient between the tyre and the road, as determining this value depends on the tyre properties.
The friction factor represents a value that can be used to estimate slipperiness of the road, and consequently the actual friction coefficient in case relevant tyre parameters are known.

\begin{figure}
\includegraphics[width=0.45\textwidth]{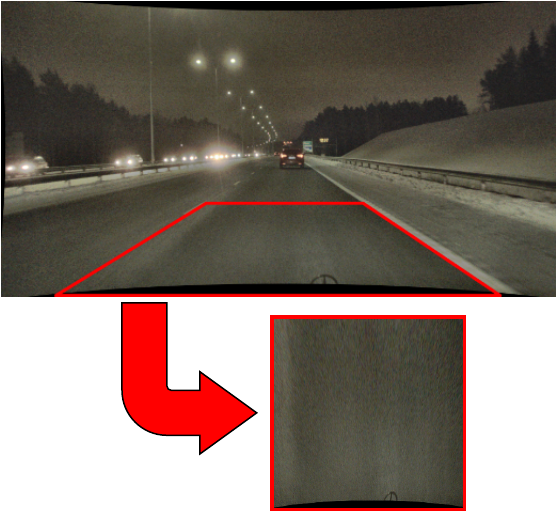}  
\caption{\small Transformation of the area in front to bird's-eye-view.}
\label{fig:bew}
\end{figure}

The images in the dataset had a resolution of 1920x1024 pixels.
To focus the analysis on the most relevant portion of the images, the road directly ahead of the vehicle, a predefined static section of all images was cropped and transformed to bird's-eye-view.
Similar pre-processing steps were utilised in \cite{cordes2022roadsaw} and \cite{vosahlik2021self}.
The cropped section was here defined to approximately represent a short section of the lane ahead of the vehicle.
A single image with clearly visible lane markings was used to find the boundaries for the cropped area.
This process is presented in Fig. \ref{fig:bew}, with the cropped area represented by a red quadrilateral in the original camera image.
The quadrilateral corresponds to the outline of the bird's-eye-view representation, as highlighted in the visualisation.
Based on knowledge of the dimensions of lane markings in Finland, in which the selected image was captured, the cropped section was inferred to represent an area with approximate dimensions of 11x4 metres. 
The consecutive bird's-eye-view transformation of the cropped section was achieved by stretching the image to a rectangle and reshaping it into a square.
The chosen boundaries for the crop operation were utilised to similarly crop each of the images in the dataset.
All further processing steps were carried out on the cropped image sections.

\begin{figure}
    \centering
    \includegraphics[width=0.45\textwidth]{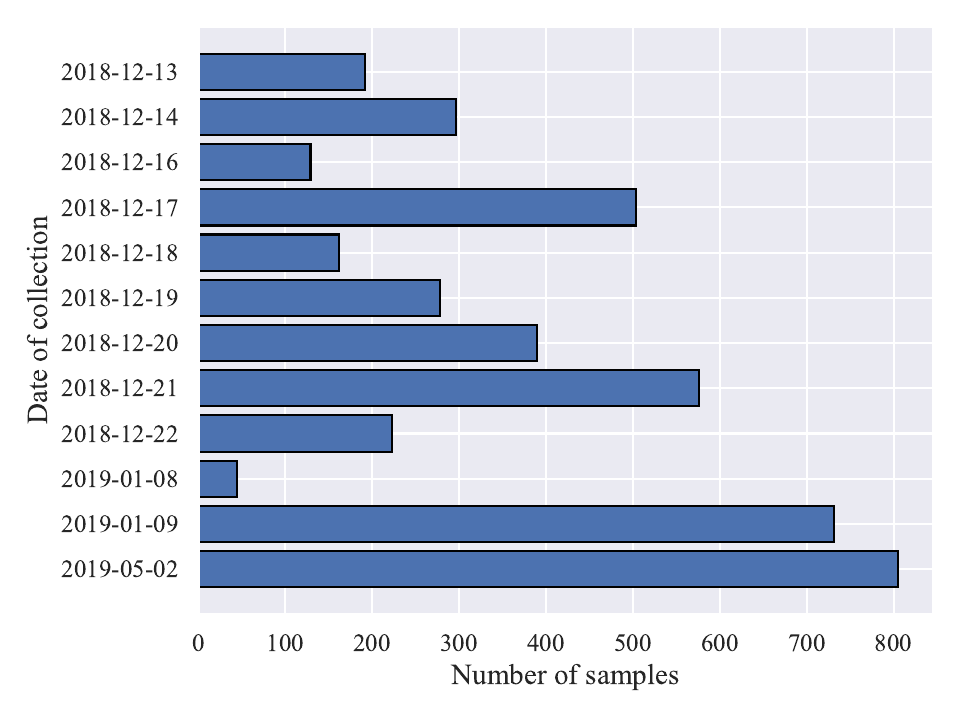}
    \caption{\small Number of samples per date in the dataset.}
    \label{fig:data_per_date}
\end{figure}

\begin{figure}
\begin{tabular}{c}
  \includegraphics[width=0.45\textwidth]{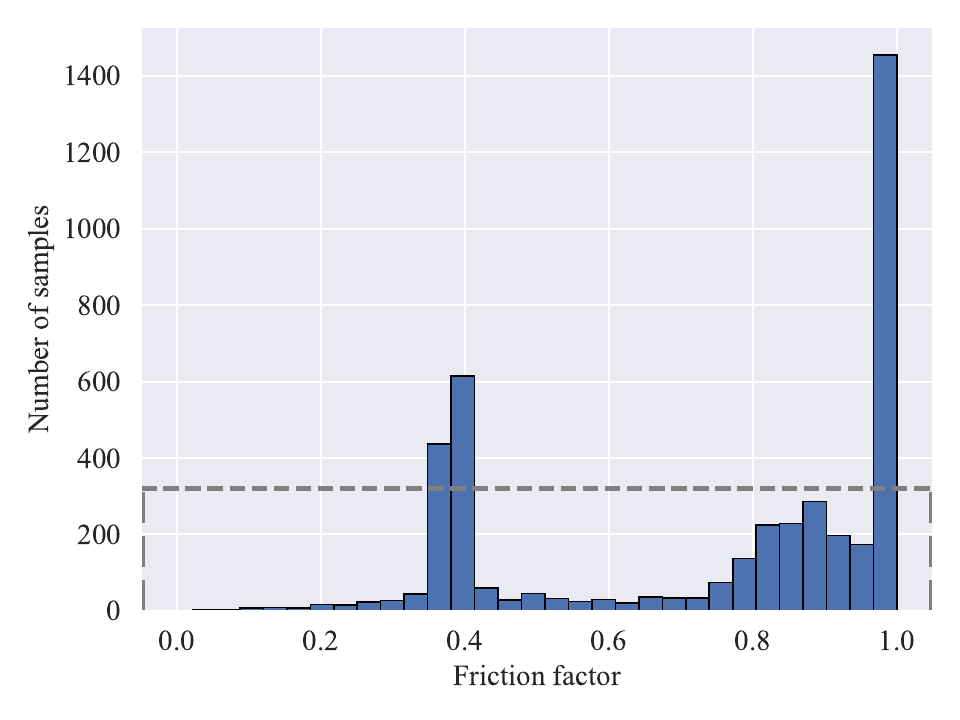} \\
  \includegraphics[width=0.45\textwidth]{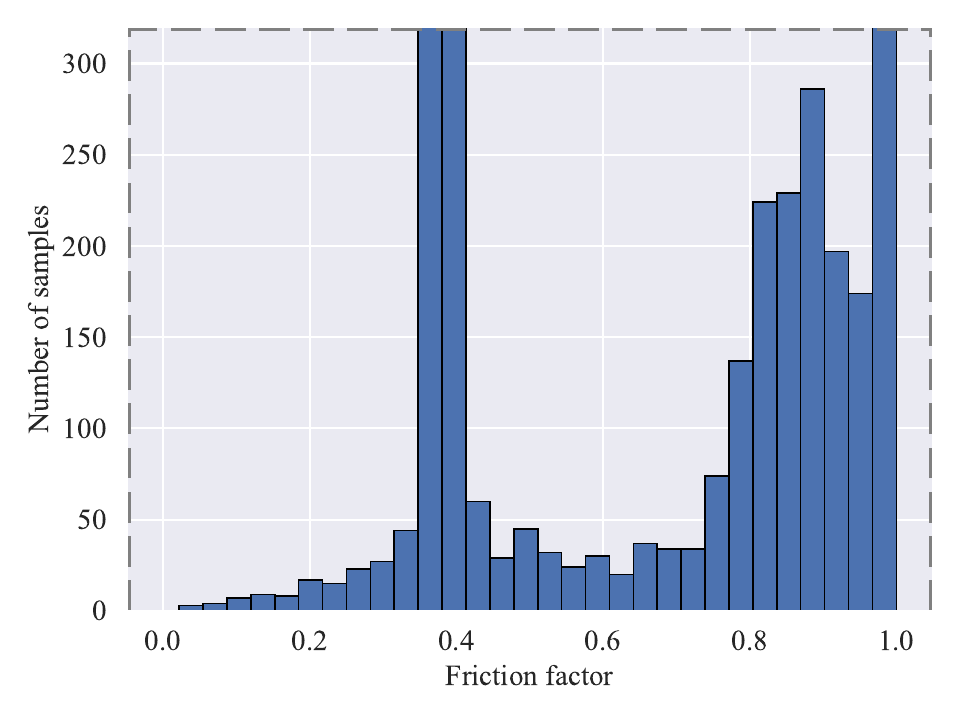}
\end{tabular}
\caption{\small Friction factor values in the dataset, with a zoomed view providing a clearer depiction of the under-represented values.}
\label{fig:friction_samples}
\end{figure}

The used dataset contained a total of 4330 samples, as this is the number of measurements in the SeeingThroughFog-dataset which contain readings from the road friction sensor.
These samples were collected in different locations on 12 different days.
Fig. \ref{fig:data_per_date} displays the number of samples collected on each date.
The distribution of the friction factor values in the dataset is visualised in Fig. \ref{fig:friction_samples}.
Image samples from the dataset with corresponding friction factor readings are provided in Fig. \ref{fig:samples}.
In addition, the overall road surface state reported by the friction sensor is presented for each sample.

{\setlength\tabcolsep{2 pt}
\begin{figure*}
\centering
\begin{tabular}{cccc}

  \includegraphics[width=0.22\textwidth]{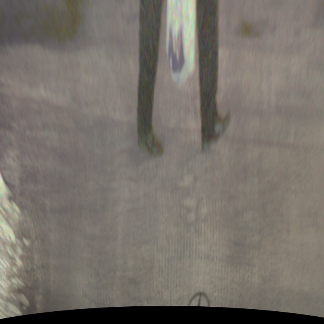} & 
   \includegraphics[width=0.22\textwidth]{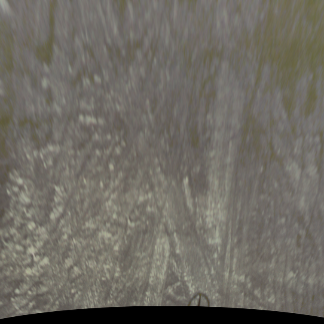} &
    \includegraphics[width=0.22\textwidth]{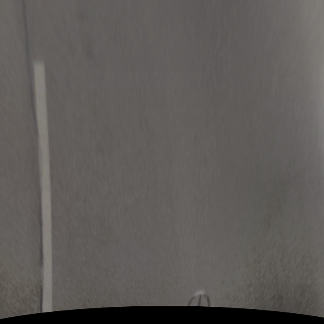} &
     \includegraphics[width=0.22\textwidth]{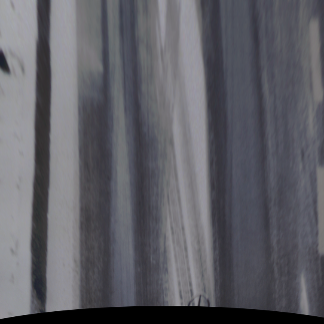} \\
 $f = 0.19$, ice & $f = 0.38$, snow & $f = 1.0$, dry & $f = 0.67$, wet \\
 \includegraphics[width=0.22\textwidth]{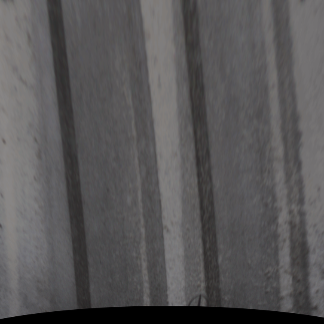} &
  \includegraphics[width=0.22\textwidth]{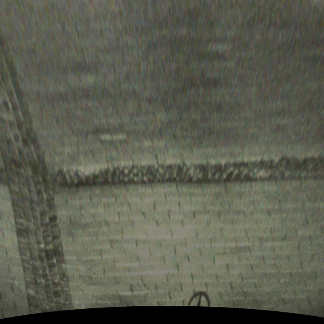} &
   \includegraphics[width=0.22\textwidth]{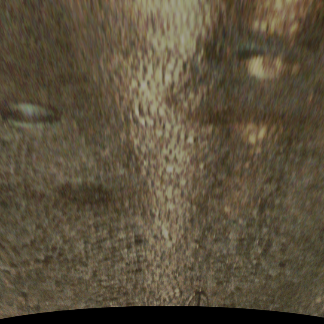} &
    \includegraphics[width=0.22\textwidth]{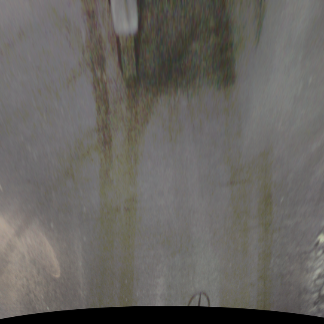} \\
 $f = 0.69$, slush & $f = 1.0$, dry & $f = 0.94$, wet & $f = 0.50$, snow \\
 \includegraphics[width=0.22\textwidth]{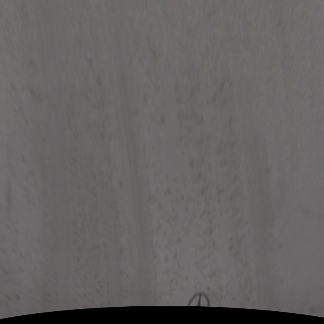} &
  \includegraphics[width=0.22\textwidth]{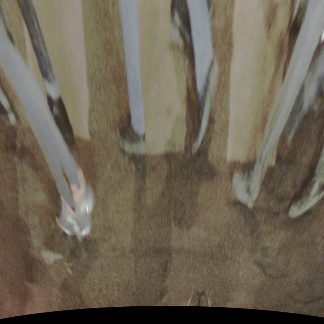} &
   \includegraphics[width=0.22\textwidth]{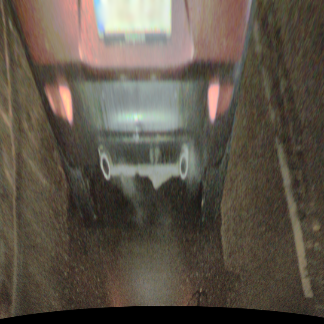} &
    \includegraphics[width=0.22\textwidth]{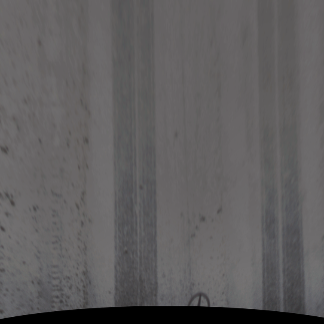} \\
 $f = 0.38$, snow & $f = 0.98$, wet & $f = 0.81$, wet & $f = 0.39$, snow 
 
\end{tabular}
\caption{Samples from the utilised data, featuring images, summary of road surface state, and corresponding ground truth friction factor values.}
\label{fig:samples}
\end{figure*}
}

\subsection{Model architecture}
\begin{figure*}
    \centering
    \includegraphics[width=0.98\textwidth]{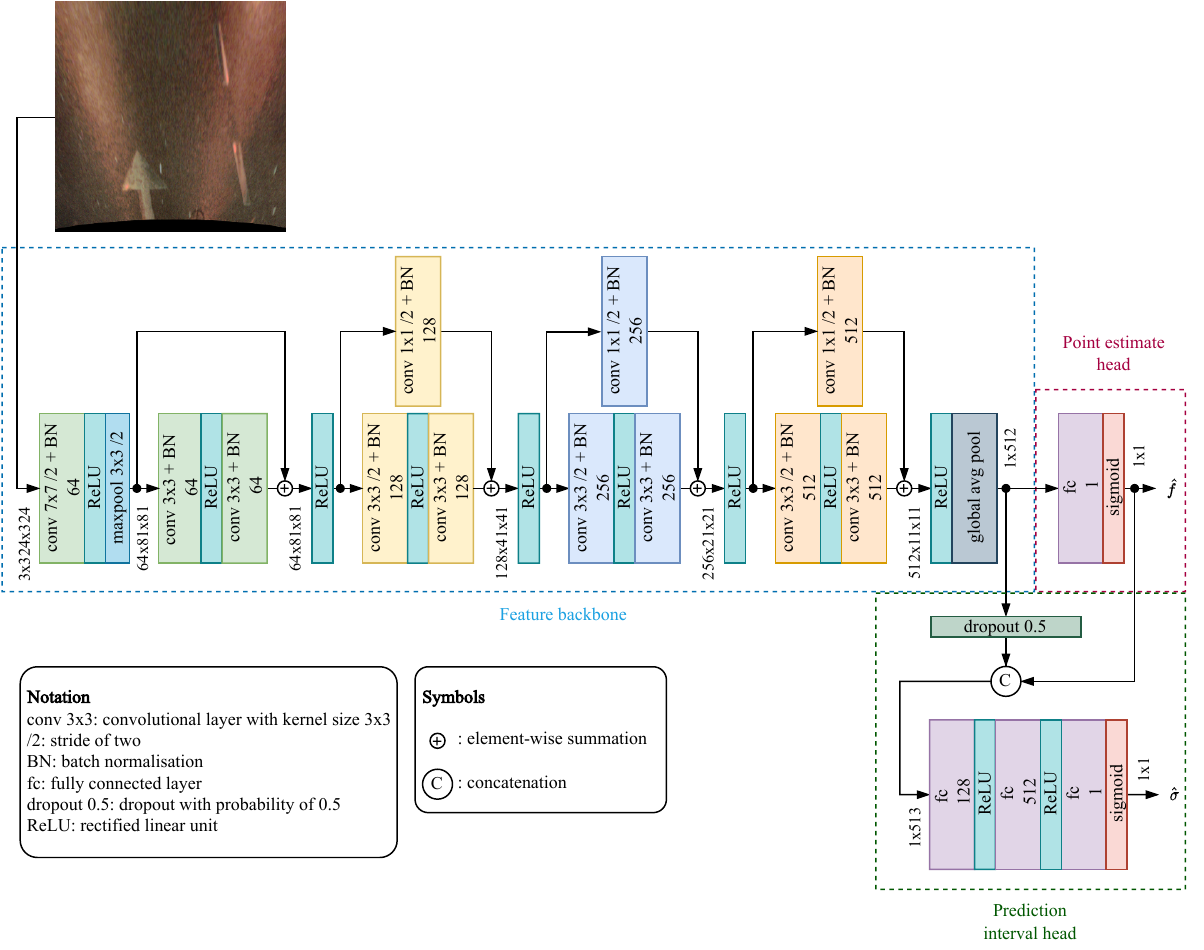}
    \caption{\small SIWNet architecture, with tensor sizes reported for processing a single image of 324x324 pixels. Each block shows the number of features in the output.}
    \label{fig:net}
\end{figure*}

The SIWNet model was developed with the main goal of presenting a lightweight model featuring uncertainty quantification for the friction factor prediction task.
The architecture of SIWNet consists of a feature backbone, as well as a point estimate head and a prediction interval head.
SIWNet architecture is presented in detail in Fig. \ref{fig:net}.
The feature backbone is responsible for processing relevant features from the images.
Based on this information, the point estimate head outputs the predicted friction factor $\hat{f}$.
The prediction interval head is responsible for assessing the uncertainty related to the point estimate.
Based on the features and the point estimate, the prediction interval head outputs a predicted standard deviation $\hat{\sigma}$, which is utilised to establish a prediction interval around the point estimate.

The feature backbone is based on the ResNet \cite{he2016deep} architecture, applying the same basic residual building blocks.
Contrary to typical ResNet implementations, each block is used only once.
This design resulted in an extremely lightweight architecture, facilitating practical applicability in embedded on-board systems.
The residual blocks were utilised as the ResNet models are widely known for their excellent accuracy and generalisablity.

In the point estimate head, a single fully connected layer is applied, which is typical for regression tasks.
A sigmoid activation function was added to the final output, as the friction factor values were bound between 0 and 1.
The prediction interval head is also constructed of fully connected layers. 
These layers analyse the features provided by the feature backbone, as well as the prediction provided by the point estimate head.
The prediction interval head analyses the friction factor prediction process, and quantifies the related uncertainty.
At the end of the prediction interval head, a sigmoid activation was added to enhance the stability of the output.
This conveniently limited the predicted standard deviation $\hat{\sigma}$ to positive numbers.
The upper limit of 1 for $\hat{\sigma}$ was deemed reasonable, since with the probabilistic model applied to train the network, this corresponded to a nearly uniform distribution within the range 0...1.

\subsection{Loss function for prediction interval head}
The prediction interval head of SIWNet is trained to quantify the uncertainty related to the prediction of the point estimate head.
Based on the quantified uncertainty, a prediction interval can be generated.
In order to perform this task, the output of SIWNet is modelled as a truncated normal distribution.
The goal of the training process is to maximise the likelihood of the training labels with regard to the predicted distributions.
Previous work has applied similar methodology with a regular normal distribution \cite{nix1994}.
A truncated normal distribution  of a random variable $x$ has a probability density function (PDF) of the form \cite{robert1995simulation}
\begin{align}
    p(\mu,\sigma,a,b;x) = \frac{\phi(\frac{x - \mu}{\sigma})} {\Phi(\frac{b - \mu}{\sigma}) - \Phi(\frac{a - \mu}{\sigma})}
\end{align}
where $\mu$ and $\sigma$ denote the mean and standard deviation of the underlying normal distribution, respectively.
The lower and upper truncation bounds are denoted by $a$ and $b$, respectively.
The truncated normal distribution PDF representation is based on the underlying normal distribution PDF, defined as
\begin{align}
    \phi(\frac{x - \mu}{\sigma}) = \frac{1}{\sigma\sqrt{2\pi}} e^{-\frac{(x-\mu)^2}{2\sigma^2}}
\end{align}
as well as the underlying normal distribution cumulative distribution function, defined as
\begin{align}
    \Phi(\frac{b - \mu}{\sigma}) = \frac{1}{2}(1 + erf(\frac{b - \mu}{\sigma\sqrt{2\pi}})),
\end{align}
where $erf(\cdot)$ is the error function.
Consequently, the negative log-likelihood of the truncated normal distribution has the form
\begin{multline}
   -\ln{p(\mu,\sigma,a,b;x)} = \ln \sigma + \frac{(\mu - x)^2}{2\sigma^2} \\ 
    + \ln{(erf(\frac{\mu - b}{\sigma \sqrt 2}) - erf(\frac{\mu - a}{\sigma \sqrt 2}))}.
\end{multline}

The training process of the prediction interval head of SIWNet is based on maximising the likelihood of the corresponding ground truths and predicted truncated normal distributions.
Consequently, this means minimising the negative log-likelihood with the predictions.
The utilised loss function for a training batch has the form
\begin{align}
    \mathcal{L} = \sum_{i=i}^n -\ln{p(\hat{f}_i,\hat{\sigma}_i,a,b;f_i)}
    \label{eq:loss}
\end{align}
where $n$ denotes the number of samples in the batch and $f$ denotes the ground truth friction factor.
The underlying normal distribution mean is the predicted friction factor $\hat{f}$ from the point estimate head, whereas the predicted standard deviation $\hat{\sigma}$ is the output of the prediction interval head.
To improve stability, $\hat{\sigma}$ is thresholded to a minimum value of $1 \cdot 10^{-4}$ when computing the loss.

\subsection{Training and testing}
For training SIWNet and conducting the presented experiments, the following steps were taken.
The utilised dataset was randomly divided into train-validation-test sets with a respective 70\%-15\%-15\% split.
When splitting the data, timestamps were utilised to ensure that data samples gathered from the same location were not included in different sets.
Hyperparameters were optimised by training the network on the training set and finding the best result on the validation set.
For the evaluation on the test set, the model was re-trained on a combination set containing both the training and validation sets. 

During the training, the feature backbone and point estimate head were first trained for 60 epochs utilising regular mean squared error as the loss function.
After these parts of the network were trained, their weights were frozen.
Afterwards, the prediction interval head was trained for 60 epochs with the loss function presented in Equation \ref{eq:loss}.
Dropout with a probability of 0.5 was applied when feeding the feature backbone output to the prediction interval head during training.

Training was carried out with a batch size of 32, and stochastic gradient descent with a momentum of 0.9 was used as the optimisation method.
During training, the learning rate was decayed with a step-based scheduler.
Every twenty epochs the learning rate was reduced to one tenth of the previous value.
As for data augmentation during training, the images were randomly flipped horizontally as well as rotated with a value from [-4,4] degrees.
Furthermore, the pixel values were slightly scaled by random color jitter in the range [0.9, 1.1].
For both training and testing, images were reshaped to 324x324 pixels before being fed to the network.
Additionally, the pixel values were normalised with the mean and standard deviation of the pixel values in the training set.

The presented experiments demonstrate comparison of SIWNet to regression models applied in the previous literature.
ResNet50 \cite{he2016deep} has been previously applied for winter condition road friction estimation \cite{vosahlik2021self}, whereas ResNet50v2 \cite{he2016identity}, VGG19 \cite{simonyan2014very} and EfficientNet-B0 \cite{tan2019efficientnet} have been applied in summer conditions \cite{du2023pavement}.
The same training and testing procedures used with SIWNet were applied with these models, except for steps related to the prediction interval head, which the other models do not include.
A sigmoid activation was also added to the output of the these models, since this was noted to boost performance.
All models were implemented on the PyTorch deep learning framework \cite{paszke2019pytorch}.
Number of trainable parameters in the models as well as floating point operations (FLOPs) executed during inference were analysed with the ptflops-tool \cite{ptflops}.

In order to benchmark the prediction intervals and the probabilistic properties of the predictions in the experiments, interval score and continuous ranked probability score (CRPS) were utilised \cite{gneiting2007strictly}.
The 90\% prediction interval was used for defining the interval score.
For constructing the prediction interval based on the truncated normal distribution output of SIWNet, the 90\% interval of the distribution with the highest likelihood was used.
Since the compared models produce only point estimates, there was no straightforward approach to define interval scores for the models.
Thus, the compared models were given static prediction intervals surrounding their point estimates.
The boundaries were set at a distance from the point estimate, which corresponded to the 90\% error threshold $e_{90\%}$ of the model on the validation set.
The 90\% error was defined as the value below which 90\% of absolute errors on the validation set were located.
The prediction interval boundaries were clamped by the plausible friction factor values, 0 to 1.

\section{Results}
SIWNet was evaluated with several experiments, to assess the efficacy of the model in the task of predicting road surface friction properties.
The computational demands of SIWNet were also analysed to determine the applicability of the model in embedded on-board usecases.
The accuracy of SIWNet was evaluated on the test set in terms of point estimates, as well as prediction intervals.
Samples of SIWNet predictions plotted next to corresponding test set images are presented in Fig. \ref{fig:test_samples}.
Each prediction plot features the point estimate of the friction factor $\hat f$, as well as the truncated normal distribution and prediction interval estimated by the prediction interval head.
Results of SIWNet were compared to those of other models, which have been applied for similar regression tasks in previous literature.

{\setlength\tabcolsep{2 pt}
\begin{figure*}
\centering
\begin{tabular}{cc}

 \includegraphics[height=0.22\textheight]{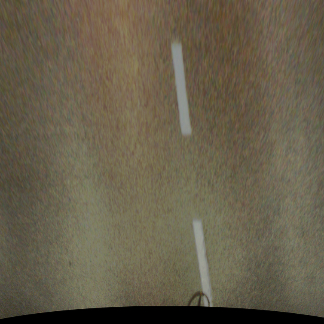} & 
  \includegraphics[height=0.22\textheight]{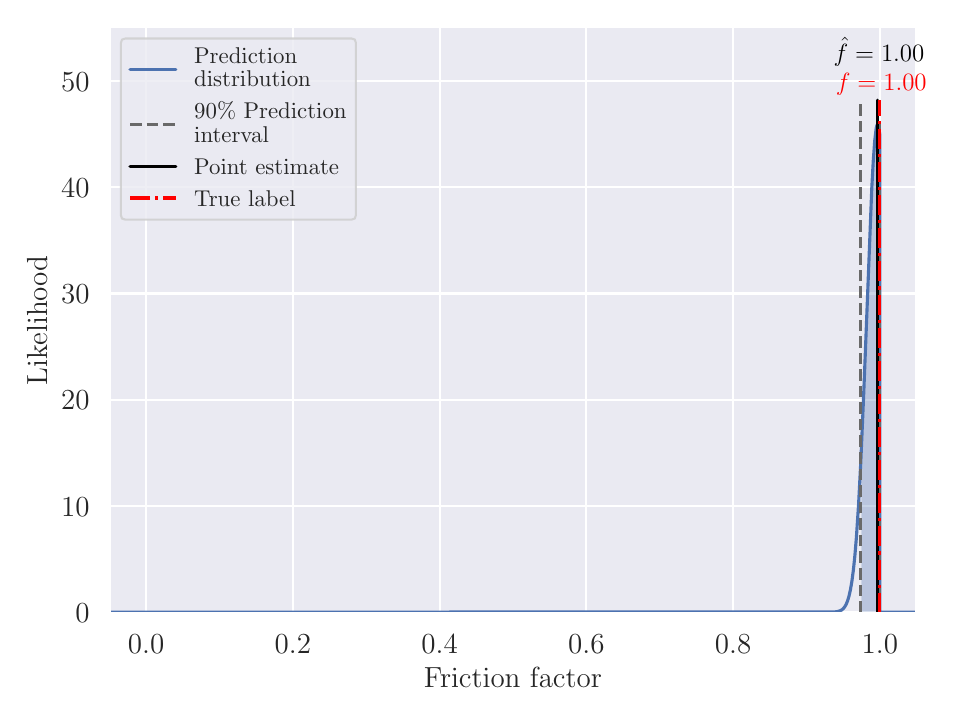} \\
 $f = 1.00$, dry & \\

 \includegraphics[height=0.22\textheight]{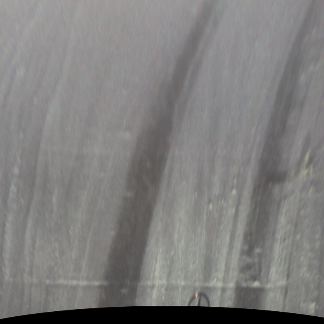} & 
  \includegraphics[height=0.22\textheight]{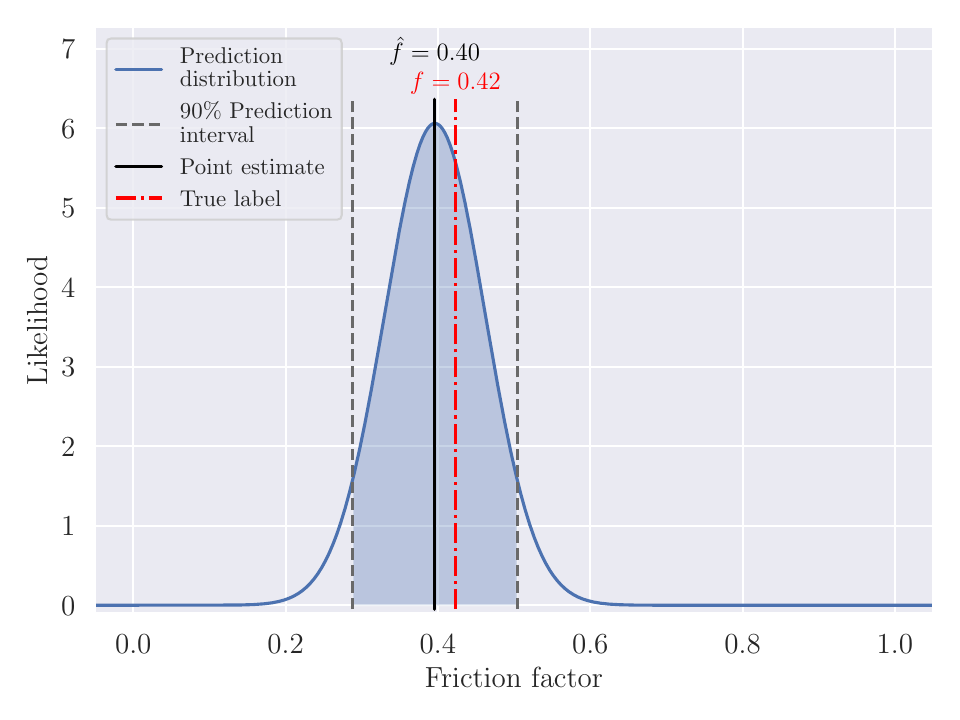} \\
 $f = 0.42$, snow & \\
 
 \includegraphics[height=0.22\textheight]{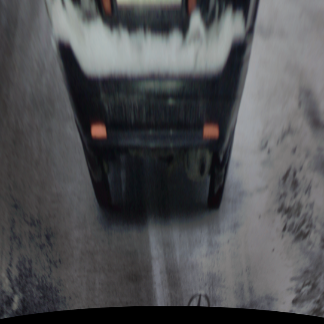} & 
  \includegraphics[height=0.22\textheight]{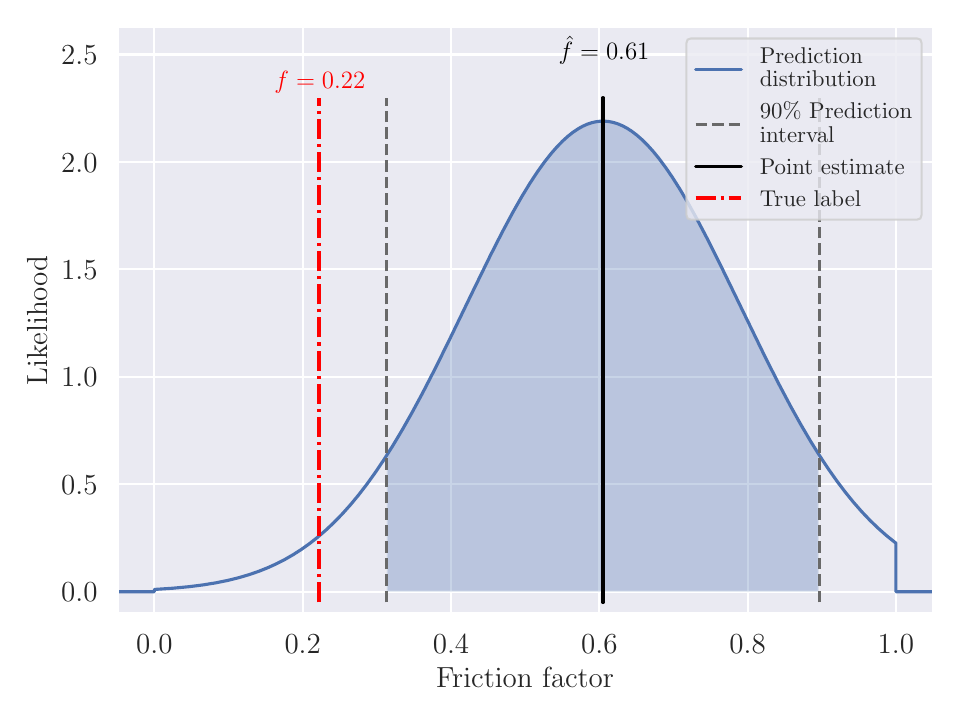} \\
 $f = 0.22$, slush & \\

 \includegraphics[height=0.22\textheight]{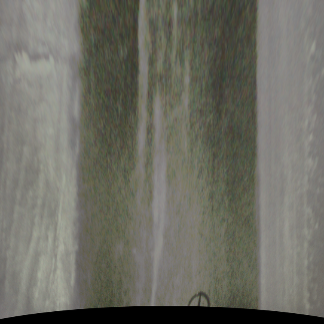} & 
  \includegraphics[height=0.22\textheight]{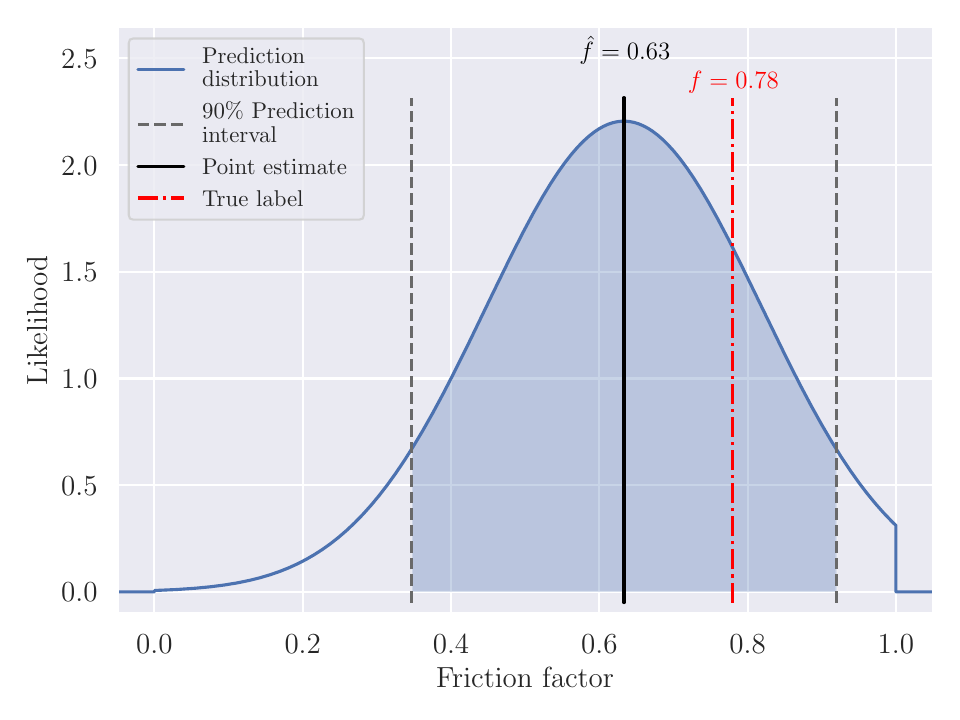} \\
 $f = 0.78$, ice & \\

\end{tabular}
\caption{SIWNET predictions on the test set with corresponding image samples, ground truth friction factor values, and road surface state.}
\label{fig:test_samples}
\end{figure*}

\subsection{Model size and computational load}
The neural network models were evaluated in terms of computational load by analysing the number of trainable parameters in the architectures, as well as the FLOPs required to perform inference on a single input image.
The SIWNet architecture was designed to require minimal computational resources, as evident from the results presented in Table \ref{tab:model_sizes}. 

\begin{table}
\caption{Model sizes and computational loads.}
\label{tab:model_sizes}
\centering
\begin{tabular}{C{0.4\columnwidth} C{0.25\columnwidth} C{0.25\columnwidth}}
\\
\noalign{\hrule height 1.2pt}
  Method & \makecell{Number of \\ parameters ($\cdot 10^6$)} & \makecell{Inference load \\ (GFLOPs)} \\
  \hline
  SIWNet (ours) & \textbf{5.04} & 4.23 \\
  EfficientNet-B0 \cite{tan2019efficientnet, du2023pavement} & 5.29 & \textbf{1.82} \\
  ResNet50v2 \cite{he2016identity, du2023pavement} & 23.5 & 18.2 \\
  ResNet50 \cite{he2016deep, vosahlik2021self} & 23.5 & 18.3 \\
  VGG19 \cite{simonyan2014very, du2023pavement} & 144 & 81.2 \\
\noalign{\hrule height 1.2pt}
\end{tabular}
\end{table}
}

Overall, SIWNet featured the lowest number of parameters of all compared models, yet utilised more FLOPs during inference than EfficientNet-B0.
Compared to ResNet50, ResNet50v2, and VGG19, SIWNet featured several times fewer parameters.
Additionally, compared to these models SIWNet also required notably fewer FLOPs to process an image.

\subsection{Point estimate head performance}
Training the feature backbone and point estimate head was the initial step for training SIWNet.
Hyperparameters were tuned with grid search, resulting in the initial learning rate and weight decay values of 0.1 and $1 \cdot 10^{-3}$, respectively.
Identical hyperparameter tuning was performed on the compared models.
Table \ref{tab:point_estimate} reports the point estimate accuracies achieved by the networks when predicting friction factors on the test data.
Accuracies are reported in terms of mean absolute error (MAE) and root-mean-square error (RMSE).

\begin{table}
\caption{Point estimate errors on the test set.}
\label{tab:point_estimate}
\centering
\begin{tabular}{C{0.4\columnwidth} C{0.25\columnwidth} C{0.25\columnwidth}}
\\
\noalign{\hrule height 1.2pt}
  Method & MAE & RMSE \\
  \hline
  SIWNet (ours) & \textbf{0.078} & \textbf{0.132} \\
  EfficientNet-B0 \cite{tan2019efficientnet, du2023pavement} & 0.100 & 0.160 \\
  ResNet50v2 \cite{he2016identity, du2023pavement} & \textbf{0.078} & 0.137 \\
  ResNet50 \cite{he2016deep, vosahlik2021self} & 0.079 & \textbf{0.132} \\
  VGG19 \cite{simonyan2014very, du2023pavement} & 0.082 & 0.144 \\
\noalign{\hrule height 1.2pt}
\end{tabular}
\end{table}

The presented results highlight that SIWNet achieved a nearly identical point estimate accuracy as ResNet50 and ResNet50v2.
VGG19 performed slightly worse, whereas EfficientNet-B0 was clearly the least accurate model.
Considering the minimal size of the SIWNet architecture, the model clearly demonstrated a favourable combination of accuracy and computational load.

\subsection{Prediction interval head performance}
The key feature of SIWNet is its capability of assessing the uncertainty of the friction factor prediction, enabled by the prediction interval head.
The prediction interval head was trained while keeping the feature backbone and point estimate head frozen.
Hyperparameters were optimised with grid search, tuning the initial learning rate, weight decay, as well as number of neurons in the fully connected layers of the prediction head.
Based on the optimisation, the initial learning rate and weight decay were set at values of $5 \cdot 10^{-4}$ and $1 \cdot 10^{-3}$, respectively.
Acquired average interval score results are presented in Table \ref{tab:pi_estimate} for SIWNet and the compared models.
As described previously, the prediction intervals of the compared models were formulated based on the 90\% error on the validation set.

\begin{table}
\caption{Average interval scores of the networks on the test set.}
\label{tab:pi_estimate}
\centering
\begin{tabular}{C{0.55\columnwidth} C{0.35\columnwidth}}
\\
\noalign{\hrule height 1.2pt}
  Method & *Average interval score \\
  \hline
  SIWNet (ours) & \textbf{0.482} \\
  EfficientNet-B0 \cite{tan2019efficientnet, du2023pavement} with $e_{90\%}$ & 0.826 \\
  ResNet50v2 \cite{he2016identity, du2023pavement} with $e_{90\%}$ & 0.722 \\
  ResNet50 \cite{he2016deep, vosahlik2021self} with $e_{90\%}$ & 0.677 \\
  VGG19 \cite{simonyan2014very, du2023pavement} with $e_{90\%}$ & 0.780 \\
\noalign{\hrule height 1.2pt}
\multicolumn{2}{c}{\footnotesize *Lower is better}
\end{tabular}
\end{table}

Based on the comparison, SIWNet clearly achieved a more favourable average interval score.
This indicates that the prediction interval head was capable of assessing the uncertainty related to the friction factor predictions of the point estimate head.

As another test to evaluate the performance of the prediction interval head, the CRPS metric was used to assess the quality of the distributions predicted by SIWNet.
The point estimates of the other detectors were also scored on the metric for comparison.
For point estimates, the average CRPS is equal to the MAE score.

\begin{table}
\caption{Average CRPS values of the networks on the test set.}
\label{tab:crps}
\centering
\begin{tabular}{C{0.45\columnwidth} C{0.45\columnwidth}}
\\
\noalign{\hrule height 1.2pt}
  Method & *Average CRPS \\
  \hline
  SIWNet (ours) & \textbf{0.060} \\
  EfficientNet-B0 \cite{tan2019efficientnet, du2023pavement} & 0.100 \\
  ResNet50v2 \cite{he2016identity, du2023pavement} & 0.078 \\
  ResNet50 \cite{he2016deep, vosahlik2021self} & 0.079 \\
  VGG19 \cite{simonyan2014very, du2023pavement} & 0.082 \\
\noalign{\hrule height 1.2pt}
\multicolumn{2}{c}{\footnotesize *Lower is better}
\end{tabular}
\end{table}

The acquired results highlight that the probabilistic forecasts of SIWNet were more representative predictions than the point estimates of the other detectors.
This indicates that the prediction interval head of SIWNet did learn a functional strategy for assessing the uncertainty of the friction factor estimates.

\subsection{Ablation study}
In order to ensure that the design decisions behind the SIWNet architecture were sensible and beneficial for the performance, an ablation study was carried out.
Similarly to the previously presented results, the ablation study analysed the point estimate head and prediction interval head separately.

The point estimate head was modified by removing the sigmoid activation, and the related training and testing procedures were repeated.
Identical actions were taken with the compared models, removing the sigmoid activations from their outputs.
During testing, the outputs of the networks were clamped between 0 and 1.
Resulting accuracies on the test set are presented in Table \ref{tab:abl_point_estimate}.
Based on the results, it is evident that the sigmoid activation was a beneficial addition for all networks.

\begin{table}
\caption{Ablation study of the point estimates.}
\label{tab:abl_point_estimate}
\centering
\begin{tabular}{C{0.55\columnwidth} C{0.15\columnwidth} C{0.15\columnwidth}}
\\
\noalign{\hrule height 1.2pt}
  Method & MAE & RMSE \\
  \hline
  SIWNet (ours) w/o sigmoid & 0.099 & \textbf{0.151} \\
  EfficientNet-B0 \cite{tan2019efficientnet, du2023pavement} w/o sigmoid & 0.289 & 0.309 \\
  ResNet50v2 \cite{he2016identity, du2023pavement} w/o sigmoid & 0.110 & 0.152 \\
  ResNet50 \cite{he2016deep, vosahlik2021self} w/o sigmoid & 0.144 & 0.190 \\
  VGG19 \cite{simonyan2014very, du2023pavement} w/o sigmoid & \textbf{0.096} & 0.153 \\
\noalign{\hrule height 1.2pt}
\end{tabular}
\end{table}

Ablation study of the SIWNet prediction interval head also investigated the effect of the sigmoid activation.
The sigmoid activation was removed from the prediction interval output, and training and testing were repeated.
In another test, the prediction interval head was trained and tested without using dropout regularisation on the feature backbone output.
Additionally, the SIWNet model was studied by training the model without the proposed technique of pretraining the feature and backbone and point estimate head, and freezing their weights.
Instead the whole network was optimised simultaneously utilising the loss function presented in Equation \ref{eq:loss}.
Finally, the overall efficacy of the prediction interval head was evaluated by formulating a static prediction interval around the point estimates based on the 90\% error on the validation set, similarly to the prediction intervals used for the predictions of the compared models.
The ablations were evaluated with the interval score, and results from the tests are presented in Table \ref{tab:abl_pi_estimate}.
All tests concluded that the originally proposed SIWNet architecture was capable of scoring higher interval scores.

\begin{table}
\caption{Ablation study of the prediction interval head.}
\label{tab:abl_pi_estimate}
\centering
\begin{tabular}{C{0.45\columnwidth} C{0.45\columnwidth}}
\\
\noalign{\hrule height 1.2pt}
  Method & *Average interval score \\
  \hline
  SIWNet w/o dropout & \textbf{0.621} \\
  SIWNet with $e_{90\%}$ & 0.710 \\
  SIWNet w/o pretrain \& freeze & 0.748 \\
  SIWNet w/o PI Sigmoid & 0.904 \\
\noalign{\hrule height 1.2pt}
\multicolumn{2}{c}{\footnotesize *Lower is better}
\end{tabular}
\end{table}

\section{Discussion}
SIWNet was capable of producing point estimates with equivalent accuracy as ResNet50 and ResNet50v2, and more accurately than the other compared models.
This is an impressive result, considering that the model has approximately 79\% fewer parameters than the ResNet models, and required roughly 77\% fewer FLOPs to process an input image.
This indicates that SIWNet offers a favourable combination of speed and accuracy, being well-fitted for on-board utilisation with limited embedded hardware.
The presented experiments also highlight that SIWNet was capable of successfully assessing the uncertainty related to its friction factor predictions.
This was evident in the CPRS as well as interval score results in Tables \ref{tab:crps} and \ref{tab:pi_estimate}.
However, exact quantification of the prediction interval capabilities of SIWNet is difficult, since only a naive method was used for comparison.
The static prediction intervals applied with the point estimates of the other methods was not an optimal strategy for uncertainty quantification.
However, since no previous works applying uncertainty quantification to road surface friction regression exist, there were no clear alternatives for comparison.
The implementation of SIWNet is published as open source, and future works on the topic are encouraged to compare their approaches to SIWNet.

Concerning the training procedure of SIWNet, the chosen approach was likely not optimal.
During training of the prediction interval head with the loss function presented in Equation \ref{eq:loss}, the point estimate head as well as the feature backbone were frozen.
This was done to stabilise the learning process, as the loss function can produce quite extreme values.
The result of a direct training approach was demonstrated in the ablation study.
However, the training stability comes at the cost of not finding optimal weights in the feature backbone for the uncertainty estimation.
Finding a strategy to optimise the feature backbone accordingly might result in better uncertainty quantification and prediction interval estimates.

Regarding the validity of the presented results overall, some unreliability was likely caused by the used dataset and its characteristics.
Notably, the included summer data in the used dataset was temporally limited to a single day, resulting in limited variance in this specific part of the dataset.
The winter data was however collected on several different days, ensuring that different winter road conditions were comprehensively included in the dataset.
Since the focus of the work was in road friction estimation in winter conditions, the limitations of the summer data should not affect the implications of this study.
Overall, a larger and more varied dataset would provide more definitive results, as the models could learn richer representations from data, and the testing would provide a more thorough and decisive investigation of the prediction capabilities.
Future work should focus on expanding the amount of data available for the development of visual road surface friction estimation.
This would be a key enabler for the development of increasingly robust prediction models.
Different real world conditions and limitations, such as other road users partially blocking the visibility of the road, would be more accurately learned and addressed by the models with an increased number of training samples.
The models should have the learning capacity to address these issues, especially SIWNet with the included uncertainty quantification methodology.
In addition, the used dataset contained a clear over-representation of certain values, as visible in Fig. \ref{fig:friction_samples}.
A more balanced dataset would likely result in better learning results of the models, as an unbalanced dataset can result in bias in the models.
These factors may have contributed to most models achieving such similar point estimate accuracies.
Increased variation in the used dataset might have resulted in more notable differences in the prediction results.

Another disadvantage of the used dataset was the fact that the road friction sensor readings did not exactly correspond to the visible road in the camera view.
This can be seen in Fig. \ref{fig:sensors}, which illustrates the sensor placement.
The road friction sensor was measuring the road area directly below the bumper of the vehicle.
This specific road area was not actually visible in the camera view simultaneously, as the camera was monitoring the road slightly ahead of the bumper.
Based on camera calibration information, the distance between the sensor and the closest edge of the monitored road area was inferred as roughly 4.6 metres.
Consequently, in order to carry out the presented analysis, an assumption had to be made that the road friction properties were the same below the bumper and in the visible road area.
This issue could be mitigated by matching the sensor reading to a previously captured image, and such approach has been adopted in some previous related studies \cite{vosahlik2021self, cordes2022roadsaw}.
This type of time series analysis was not possible here, as the utilised dataset contained temporally sparse samples.
To the best of the authors' knowledge, the used dataset was the only openly available dataset featuring images and friction sensor readings in winter conditions.
However, assumption of uniform road condition below the bumper and slightly ahead in the camera view should generally hold.
This was also indicated by the point estimate accuracy results of the different networks, as the achieved prediction errors can be considered fairly low.
Classification-based studies typically cluster the road surface conditions to a single digit number of clusters, e.g. in the review of Ma \textit{et al.} \cite{ma2022current} a total number of seven clusters were identified.
Since the MAE and RMSE values scored by the networks were roughly around 10\% of the total range of the friction factor value space, the errors were on average smaller than the resolution of classification approaches in general.
Therefore, the friction sensor reading not perfectly matching the image should not have had too drastic an impact.
The importance of the sensor placement is further diminished by the fact that a single friction sensor reading was used to label an entire image, which in any case demands an assumption of uniform road surface conditions.

Since the road friction sensor measured only a small area of the road, whereas the camera image captured a large portion of the road ahead, there were clear spatial limitations in the used data.
Similar limitations can be found in previous studies in the field \cite{vosahlik2021self, cordes2022roadsaw, du2023pavement}, as generally entire images are labelled with a single ground truth value acquired from a friction sensor or derived from wheel slip.
Consequently, the ground truth only partially covers the spatial range of the source image.
In some scenarios, the visible road conditions may vary greatly within the image.
For example, in winter conditions the road may be partially covered in snow, while also featuring tyre tracks with clear asphalt or water.
This naturally leads to a certain degree of ambiguity and noise in the data, as the ground truth value does not fully represent the information in the image.
This likely affected the accuracy as well as general applicability of the models used here.
As a result, the models likely learned optimal weights which attempted to ignore as much of the noise as possible.
However, the presented results still indicate that the models were able to predict friction properties consistently, despite the noise and spatial limitations of the data.
An optimal dataset for road surface friction estimation purposes should include several ground truth labels for the visible road area, allowing more fine-grained estimation.
This could be achieved by for example mounting a vehicle with several measurement units.
To the best of the authors' knowledge, no such public dataset unfortunately exists.
Future studies are encouraged to investigate this topic with more advanced sensor setups.

It should also be noted that the utilised dataset, and consequently the trained models, only consider road friction properties related to snow, ice, and water.
In winter conditions, these factors largely determine the overall friction properties of the road, resulting in drastic differences in tyre-road friction.
This study focused on these conditions due to their critical impact on vehicle dynamics.
In summer conditions, road friction properties are most notably affected by water on the road, as well as the road surface pavement type.
Analysis of the road surface pavement type was ignored here, which slightly limits the generalisability of the results.
For improved generalisability, future development of SIWNet could aim to include the road pavement type as a factor in the analysis.

\section{Conclusion}
This work enhanced camera-based winter road condition monitoring by presenting the SIWNet model.
Based on an image of the road, the deep learning regression model predicts a friction factor, which summarises the friction properties of the visible road.
The main goal of this study was to advance previous state of the art by including uncertainty quantification in the prediction model, as well as designing the model to require minimal computational resources.
SIWNet is computationally lightweight, and features a built-in uncertainty quantification mechanism, allowing the model to generate prediction intervals instead of traditional point estimates.
Due to these characteristics, SIWNet advances robustness of road condition monitoring via computer vision.
This is a key advancement, as reliable road condition monitoring is vital for proper tuning of controllers in automated vehicle applications.

Future research efforts on the topic should focus on more fine-grained spatial analysis of winter road surfaces.
Computer vision solutions for road condition monitoring could be extended to pixel-level analysis of the road surface.
Furthermore, spatio-temporal data analysis might also provide clear benefits to the accuracy at which road condition can be monitored.
Integrating such features to the SIWNet architecture could allow for even more reliable modelling and estimation of the road surface conditions in the future.
Considering the wide variety of winter road conditions and their effect on optimal automated driving strategies, additional research efforts should be concentrated on developing increasingly robust solutions for these demanding circumstances.

In order to improve the reliability of automated driving solutions, the technology must be capable of operating in all weather and road conditions.
Proper situational awareness of vehicles in adverse conditions is a key factor in enabling safe and robust future automated vehicle solutions.
SIWNet advances the road condition monitoring capabilities of vehicles in winter conditions, potentially enhancing the operation of automated driving functionalities and bringing reliable automated driving functions closer to reality.

\section*{Acknowledgments}
The authors wish to acknowledge the funding provided by Henry Ford Foundation Finland. 

\bibliographystyle{IEEEtran}
\bibliography{refs}

\newpage

\section{Biography Section}
 
\vspace{11pt}

\vspace{-33pt}
\begin{IEEEbiography}[{\includegraphics[width=1in,height=1.25in,clip,keepaspectratio]{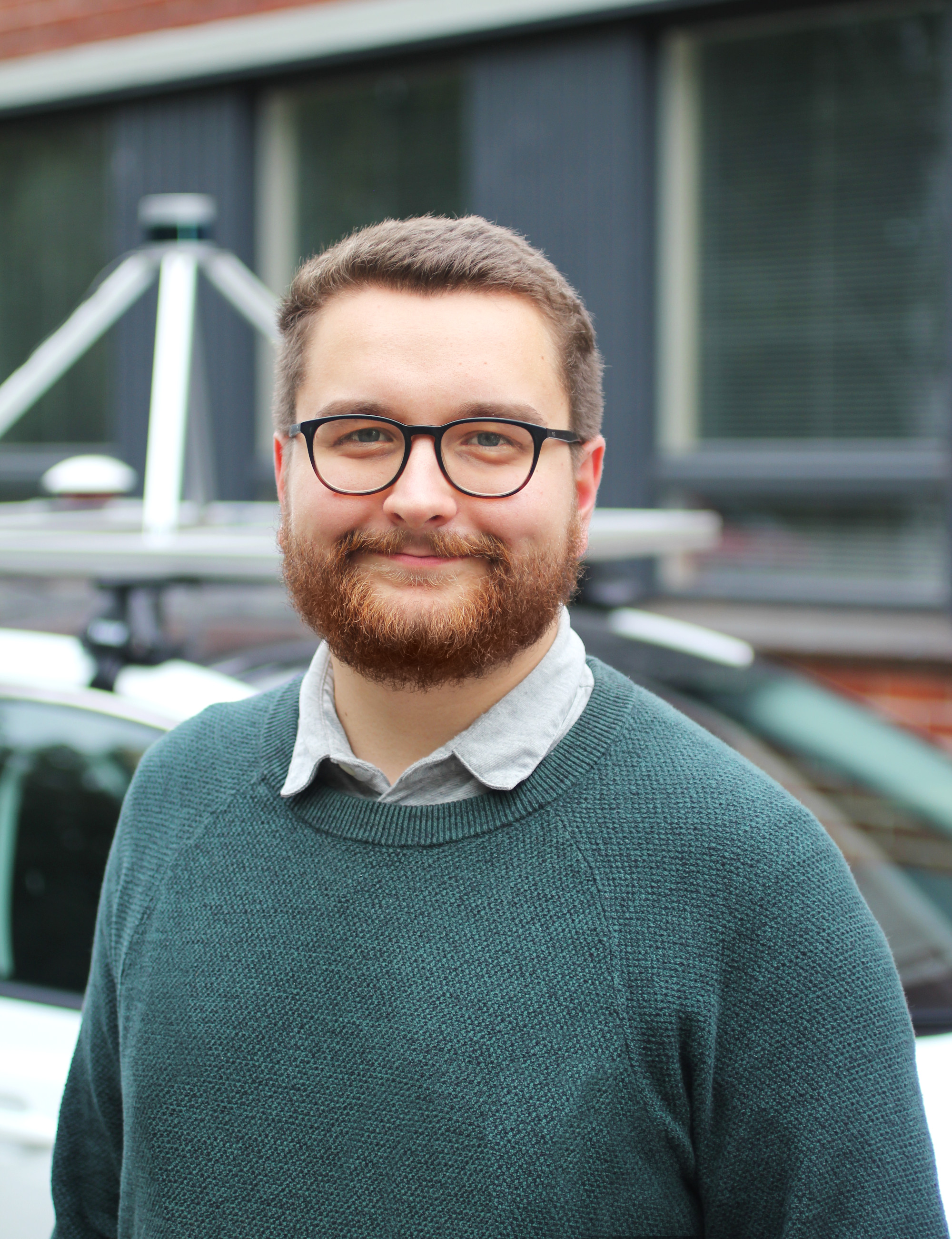}}]{Risto Ojala}
received his BSc, MSc, and DSc degrees from Aalto University
in 2019, 2021, and 2023, respectively. 
Currently, he continues at Aalto as a postdoctoral researcher at the Autonomy \& Mobility laboratory. 
He has authored several peer-reviewed journal publications, and his research interests focus on automated vehicles, mobile robotics, computer vision, and machine learning.
\end{IEEEbiography}

\vspace{11pt}

\begin{IEEEbiography}[{\includegraphics[width=1in,height=1.25in,clip,keepaspectratio]{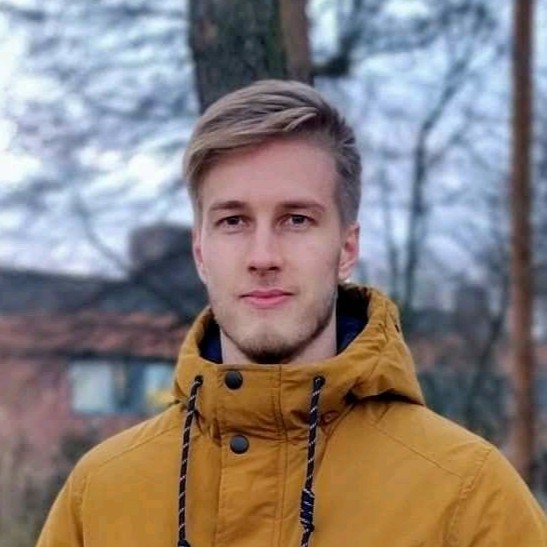}}]{Alvari Seppänen}
received his BSc and MSc degrees from Aalto University in 2020 and 2021, respectively. 
He is currently studying towards a DSc degree at Aalto University in the Autonomy \& Mobility laboratory. 
His research interests are mobile robotics and robot perception.
He has multiple peer-reviewed publications on these topics.
\end{IEEEbiography}

\vfill

\end{document}